\documentclass{article}

\usepackage{microtype}
\usepackage{graphicx}
\usepackage{subfigure}
\usepackage{booktabs} %

\usepackage[pdfa, pdfdisplaydoctitle, pdflang={en-US}]{hyperref}

\usepackage[accepted]{icml2021}

\usepackage{bm}

\usepackage[utf8]{inputenc}
\usepackage[T1]{fontenc}
\DeclareUnicodeCharacter{2212}{-}

\usepackage{xcolor}

\usepackage{tikz}
\usetikzlibrary{shapes}
\usetikzlibrary{automata,topaths}
\usetikzlibrary{calc}
\usepackage{pgfplots}
\pgfplotsset{compat=1.16}

\usepackage{listings}%
\usepackage{csvsimple}

\usepackage{wrapfig}

\usepackage{pdfpages}

\usepackage{afterpage}

\usepackage{rotating}
\usepackage{enumitem}

\usepackage{mathtools}

\usepackage{dsfont}

\usepackage{multirow}
\usepackage{colortbl}

\usepackage[toc,page]{appendix}

\usepackage{url}

\usepackage{caption}

\usepackage{amsbsy}
\usepackage{stmaryrd}

\usepackage{nicefrac}

\usepackage{amsmath,amsfonts,bm}

\def\eqref#1{equation~\ref{#1}}

\def\1{\bm{1}}

\DeclareMathAlphabet{\mathsfit}{\encodingdefault}{\sfdefault}{m}{sl}
\SetMathAlphabet{\mathsfit}{bold}{\encodingdefault}{\sfdefault}{bx}{n}

\newcommand{\sigmoid}{\sigma}

\newcounter{arrown}

\let\pgfimageWithoutPath\pgfimage 
\renewcommand{\pgfimage}[2][]{\pgfimageWithoutPath[#1]{fig/#2}}

\newcommand{\mnistimgheight}{.85em}
\newcommand{\mnistimgraise}{-.1em}
\newcommand{\mnistimg}[1]{\raisebox{\mnistimgraise}{\includegraphics[height=\mnistimgheight]{#1}}}

\newcommand{\svhnimg}[1]{\raisebox{-.4em}{\includegraphics[height=1.35em]{#1}}}

\newif\ifdraft
\drafttrue

\ifdraft
\newcommand{\fpc}[1]{{\color{cyan}\textbf{FP:} #1}}
\newcommand{\odc}[1]{{\color{orange}\textbf{OD:} #1}}
\newcommand{\rwc}[1]{{\color{green!70!black}\textbf{RW:} #1}}
\newcommand{\cbc}[1]{{\color{yellow!50!black}\textbf{CB:} #1}}
\newcommand{\hkc}[1]{{\color{purple}\textbf{HK:} #1}}

\else
\newcommand{\fpc}[1]{}
\newcommand{\odc}[1]{}
\newcommand{\rwc}[1]{}
\newcommand{\cbc}[1]{}
\newcommand{\hkc}[1]{}

\fi
\usepackage{multicol}

\icmltitlerunning{Differentiable Sorting Networks for Scalable Sorting and Ranking Supervision}
\begin{document}

\twocolumn[
\icmltitle{Differentiable Sorting Networks for Scalable Sorting and Ranking Supervision}

\begin{icmlauthorlist}
\icmlauthor{Felix Petersen}{kn}
\icmlauthor{Christian Borgelt}{salz}
\icmlauthor{Hilde Kuehne}{fra,mitibm}
\icmlauthor{Oliver Deussen}{kn}
\end{icmlauthorlist}

\icmlaffiliation{kn}{University of Konstanz, Germany}
\icmlaffiliation{salz}{University of Salzburg, Austria}
\icmlaffiliation{fra}{University of Frankfurt, Germany}
\icmlaffiliation{mitibm}{MIT-IBM Watson AI Lab}

\icmlcorrespondingauthor{Felix Petersen}{felix.petersen@uni.kn}

\icmlkeywords{Machine Learning, ICML, Sorting Networks, Differentiable Algorithm, Ranking Supervision}

\vskip 0.3in
]

\printAffiliationsAndNotice{}  %

\begin{abstract}
    
Sorting and ranking supervision is a method for training neural networks end-to-end based on ordering constraints. That is, the ground truth order of sets of samples is known, while their absolute values remain unsupervised. For that, we propose differentiable sorting networks by relaxing their pairwise conditional swap operations. To address the problems of vanishing gradients and extensive blurring that arise with larger numbers of layers, we propose mapping activations to regions with moderate gradients. We consider odd-even as well as bitonic sorting networks, which outperform existing relaxations of the sorting operation. We show that bitonic sorting networks can achieve stable training on large input sets of up to 1024 elements.

\end{abstract}

\section{Introduction}

\begin{figure*}
    \centering
    \includegraphics[width=.92\textwidth]{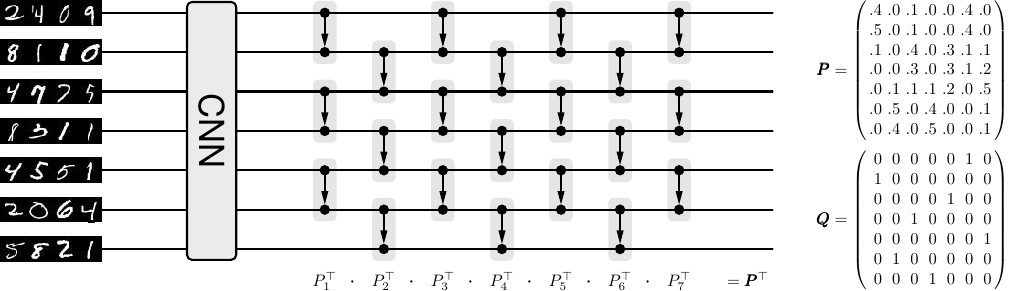}
    \caption{
    Overview of the system for training with sorting supervision. 
    Left: input images are fed separately / independently into a Convolutional Neural Network (CNN) that maps them to scalar values.
    Center: the odd-even sorting network sorts the scalars by parallel conditional swap operations.
    Right: the sorting network produces a differentiable permutation matrix $\pmb{P}$ which can then be compared to the ground truth permutation matrix $\pmb{Q}$ using binary cross-entropy to produce the training loss.
    By propagating this error backward through the sorting network, we can train the CNN.
    }
    \label{fig:overall-architecture-odd-even}
\end{figure*}

Sorting and ranking as the ability to score elements by their relevance is an essential task in numerous applications.
It can be used for choosing the best results to display by a search engine or organize data in memory.
Starting in the 1950s, sorting networks have been presented to address the sorting task \cite{Knuth1998-3-SortingSearching}.
Sorting networks are sorting algorithms with a fixed execution structure, which makes them suitable for hardware implementations, e.g., as part of circuit designs.
They are oblivious to the input, i.e., their execution structure is independent of the data to be sorted.
As such hardware implementations are significantly faster than conventional multi-purpose hardware, they are of interest for sorting in high performance computing applications \cite{Govindaraju06GPUTeraSort}.
This motivated the optimization of sorting networks toward faster networks with fewer layers, which is a still-standing problem \cite{bidlo2019evolutionary}.
Note that, although the name is similar, sorting networks are \emph{not} neural networks that perform sorting.

Recently, the idea of end-to-end training of neural networks with sorting and ranking supervision by a continuous relaxation of the sorting and ranking functions has been presented by \citet{Grover2019-NeuralSort}.
Sorting supervision means the ground truth order of some samples is known while their absolute values remain unsupervised.
As the error has to be propagated in a meaningful way back to the neural network, it is necessary to use a continuous and continuously differentiable sorting function.
Several such differentiable relaxations of the sorting and ranking functions have been introduced, e.g., by \citet{adams2011ranking}, \citet{Grover2019-NeuralSort}, \citet{Cuturi2019-SortingOT}, and \citet{Blondel2020-FastSorting}.
For example, they enable training a CNN based on ordering and ranking information instead of absolute ground truth values.
As sorting a sequence of values requires finding the respective ranking order, we use the terms ``sorting'' and ``ranking'' interchangeably.

In this work, we propose to combine traditional sorting networks and differentiable sorting functions by presenting smooth differentiable sorting networks. 

Sorting networks are conventionally non-differentiable as they use $\min$ and $\max$ operators for conditionally swapping elements.
Thus, we relax these operators by building on the $\operatorname{softmin}$ and $\operatorname{softmax}$ operators.
However, due to the nature of the sorting network, values with large as well as very small differences are compared in each layer.
Comparing values with large differences causes vanishing gradients, while comparing values with very small differences can modify, i.e., blur, values as they are only partially swapped.
This is because $\operatorname{softmin}$ and $\operatorname{softmax}$ are based on the logistic function which is saturated for large inputs but also returns a value close to the mean for inputs that are close to each other.
Based on these observations, we propose an activation replacement trick, which avoids vanishing gradients as well as blurring.
That is, we modify the distribution of the differences between compared values to avoid small differences close to $0$ as well as large differences.

To validate the proposed idea and to show its generalization, we evaluate two sorting network architectures, the odd-even as well as the bitonic sorting network.
The idea of odd-even sort is to iteratively compare adjacent elements and swap pairs that are in the wrong order. 
The method alternately compares all elements at odd and even indices with their successors. 
To make sure that the smallest (or greatest) element will be propagated to its final position for any possible input of length $n$, we need $n$ exchange layers. 
An odd-even network is displayed in Figure~\ref{fig:overall-architecture-odd-even} (center).
Odd-even networks can be seen as the most generic architectures, and are mainly suitable for small input sets as their number of layers directly depends on the number of elements to be sorted.

Bitonic sorting networks \cite{batcher1968sorting} use bitonic sequences to sort based on the Divide-and-Conquer principle and allow sorting in only $\mathcal{O}({\log^2n})$ parallel time. 
Bitonic sequences are twice monotonic sequences, i.e., they consist of a monotonically increasing and monotonically decreasing sequence.
Bitonic sorting networks recursively combine pairs of monotonic sequences into bitonic sequences and then merge them into single monotonic sequences.
Starting at single elements, they eventually end up with one sorted monotonic sequence. 
With the bitonic architecture, we can sort large numbers of input values as we only need $\frac{\log_2n \cdot ((\log_2n)+1)}{2}$ layers to sort $n$ inputs. 
As a consequence, the proposed architecture provides good accuracy even for large input sets and allows scaling up sorting and ranking supervision to large input sets of up to $1024$ elements.

Following \citet{Grover2019-NeuralSort} and \citet{Cuturi2019-SortingOT}, we benchmark our continuous relaxation of the sorting function on the four-digit MNIST \cite{LeCun2010-mnist} sorting supervision benchmark. 
To evaluate the performance in the context of a real-world application, we apply our continuous relaxation to the multi-digit images of the Street View House Number (SVHN) data set. 
We compare the performance of both sorting network architectures and evaluate their characteristics under different conditions. 
We show that both differentiable sorting network architectures outperform existing continuous relaxations of the sorting function on the four-digit MNIST sorting benchmark and also perform well on the more realistic SVHN benchmark.
Further, we show that our model scales and achieves performance gains on larger sets of ordered elements and confirm this up to $n=1024$ elements.

An overview of the overall architecture is shown in Figure~\ref{fig:overall-architecture-odd-even}.

In addition, we apply our method to top-$k$ classification.

\section{Related work}

\begin{figure*}[t]
\centering
\includegraphics[height=41mm]{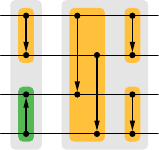}\hfill
\includegraphics[height=41mm]{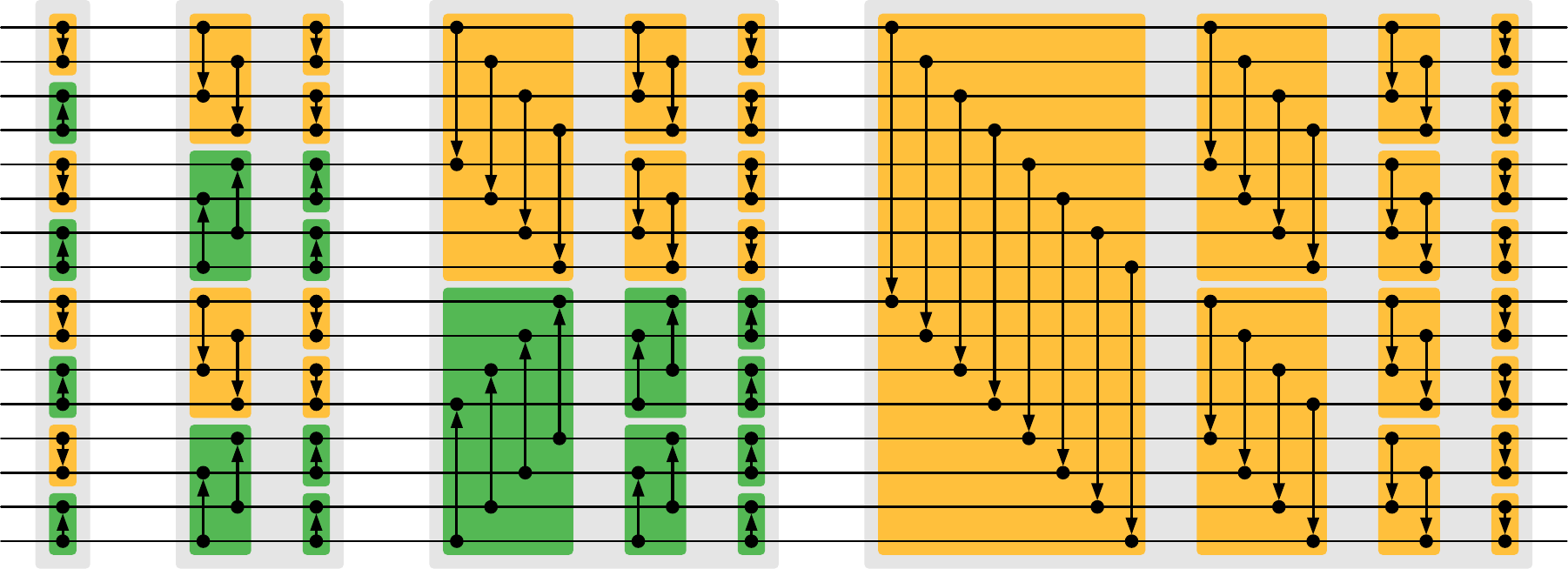}
\caption{
    Bitonic sorting networks for 4 and 16 lanes, consisting of bitonic merge blocks (colored). Arrows pointing toward the maximum.
    \label{fig:bitonic_sort} \label{fig:bitonic-schema}
    }
\end{figure*}

\paragraph{Sorting Networks} The goal of research on sorting networks is to find optimal sorting networks, i.e., networks that can sort an input of $n$ elements in as few layers of parallel swap operations as possible.
Initial attempts to sorting networks required $\mathcal{O}(n)$ layers, each of which requires $\mathcal{O}(n)$ operations (examples are bubble and insertion sort \cite{Knuth1998-3-SortingSearching}).
With parallel hardware, these sorting algorithms can be executed in $\mathcal{O}(n)$ time.
Further research lead to the discovery of the bitonic sorting network (aka.~bitonic sorter) which requires only $\mathcal{O}(\log^2n)$ layers~\cite{Knuth1998-3-SortingSearching, batcher1968sorting}.
Using genetic and evolutionary algorithms, slightly better optimal sorting networks were found for specific $n$ \cite{bidlo2019evolutionary, baddar2012designing}. 
However, these networks do not exhibit a simple, regular structure.
Ajtai, Koml\'os, and Szemer\'edi \cite{Ajtai1983-AKS} presented the AKS sorting network which can sort in $\mathcal{O}(\log n)$ parallel time, i.e., using only $\mathcal{O}(n\log n)$ operations.
However, the complexity constants for the AKS algorithm are to date unknown and optimistic approximations assume that it is faster than bitonic sort if and only if $n \gg 10^{80}$. 
Today, sorting networks are still in use, e.g., for fast sorting implementations on GPU accelerated hardware as described by \citet{Govindaraju06GPUTeraSort} and in hybrid systems as described by \citet{Gowanlock19Hybrid}. 
Based on the bitonic sorting network, \citet{lim2016box} propose a coordinate descent algorithm to solve hard permutation problems.

\paragraph{Neural Networks that Sort}
In the past, neural networks that sort have been proposed, e.g., by \citet{Ceterchi2008SpikingNP}, who proposed simulating sorting networks with spiking neural P systems.
Spiking neural P systems are predecessors of current spiking networks, a form of computational models inspired by biological neurons.
This was later adapted by \citet{MettaKelemenov15Sorting} for a spiking neural P system with anti-spikes and rules on synapses.

\citet{Graves14Neural} raised the idea of integrating sorting capabilities into neural networks in the context of Neural Turing Machines (NTM). 
The NTM architecture contains two basic components: a neural network controller based on an LSTM and a memory bank with an attention mechanism, both of which are differentiable. 
The authors use this architecture to sort sequences of binary vectors according to given priorities.
\citet{Vinyals16OrderMatters} address the problem of the order of input and output elements in LSTM sequence-to-sequence models by content-based attention.
To show the effect of the proposed model, they apply it to the task of sorting numbers and formulate the task of sorting as an instance of the set2seq problem.
\citet{Mena2018Sinkhorn} introduce the Gumbel-Sinkhorn, a Sinkhorn-operator--based analog of the Gumbel-Softmax distribution for permutations.
They evaluate the proposed approach, i.a., on the task of sorting up to $120$ numbers.
Note that these architectures learn to sort, while sorting networks and differentiable sorting functions sort provably correct.
These methods allow sorting input values, as an alternative to classical sorting algorithms, but not training with sorting supervision because they are not differentiable.

\paragraph{Differentiable Sorting}  
Closest to our work are differentiable sorting algorithms, which can be used to train neural networks based on sorting and ranking supervision.

\citet{adams2011ranking} propose relaxing permutation matrices to doubly-stochastic matrices based on marginals of distributions over permutation matrices.
They apply their method to the LETOR learning-to-rank benchmark \cite{liu2011learning}.

\citet{Grover2019-NeuralSort} propose NeuralSort, a continuous relaxation of permutation matrices to the set of unimodal row-stochastic matrices via the Plackett-Luce family of distributions over permutations.
For evaluation, they propose the benchmark of predicting the scalar value displayed on concatenated four-digit MNIST numbers. 
As supervision, they use the ranking of between $3$ and $15$ of those numbers.
Additionally, they apply NeuralSort to differentiable quantile regression and $k$-nearest neighbors image classification.

Following this work, \citet{Cuturi2019-SortingOT} presented a method for smoothed ranking and sorting operators using optimal transport (OT).
They use the idea that sorting can be achieved by minimizing the matching cost between elements and an auxiliary target of increasing values.
That is, the smallest element is matched to the first value, the second smallest to the second value, etc.
They make this differentiable by regularizing the OT problem with an entropic penalty and solving it by applying Sinkhorn iterations.
Additionally, they devise a differentiable top-$k$ operator for top-$k$ supervised image classification.
Based on this idea, \citet{xie2020differentiable} have used OT and the differentiable top-$k$ operator for $k$-nearest neighbors image classification and differentiable beam search.

Recently, \citet{Blondel2020-FastSorting} presented the idea of constructing differentiable sorting and ranking operators as projections onto the permutahedron, the convex hull of permutation matrices.
They solve this by reducing it to isotonic optimization and make it differentiable by considering the Jacobians of the isotonic optimization and the projection.
They apply their method to top-$k$ supervised image classification, label ranking via a differentiable Spearman's rank correlation coefficient, and robust regression via differentiable least trimmed squares.

\section{Sorting Networks}

In this section, we introduce two common sorting networks: the simple odd-even sorting network as well as the more complex, but also more efficient, bitonic sorting network.

\subsection{Odd-Even Sorting Network}

One of the simplest sorting networks is the fully connected odd-even sorting network.
Here, neighboring elements are swapped if they are in the wrong order.
As the name implies, this is done in a fashion alternating between comparing odd and even indexed elements with their successors.
In detail, for sorting an input sequence $a_1a_2...a_n$, each layer updates the elements such that $a'_i = \min(a_i, a_{i+1})$ and $a'_{i+1} = \max(a_i, a_{i+1})$ for all odd or even indices $i$, respectively.
Using $n$ of such layers, a sequence of $n$ elements is sorted as displayed in Figure~\ref{fig:overall-architecture-odd-even}~(center).

\subsection{Bitonic Sorting Network}

Second, we review the bitonic sorting network for sorting $n=2^k$ elements where $k\in\mathbb{N}_+$.
If desired, the sorting network can be extended to $n\in \mathbb{N}_+$ \cite{Knuth1998-3-SortingSearching}. 

The bitonic sorting networks builds on bitonic sequences: a sequence $(a_i)_{1\leq i < n}$ is called bitonic if (after an appropriate circular shift) $a_1\leq...\leq a_j \geq...\geq a_{n}$ for some $j$. %

Following the Divide-and-Conquer principle, in analogy to merge sort, bitonic sort recursively splits the task of sorting a sequence into the tasks of sorting two subsequences of equal length, which are then combined into a bitonic sequence.
Like merge sort, bitonic sort starts by merging individual elements, to obtain sorted lists of length~$2$ (first gray block in Figure~\ref{fig:bitonic_sort}).
Pairs of these are then combined into bitonic sequences and then merged into monotonic sequences (second gray block in Figure~\ref{fig:bitonic_sort}). %
This proceeds, doubling the length of the sorted sequences with each (gray) block until the entire sequence is sorted. 
The difference to merge sort lies in the bitonic merge operation, which merges two sequences sorted in opposite order (i.e., a single bitonic sequence) into a single sorted (monotonic) sequence.

In Supplementary Material~\ref{sm:bitonic}, we give more details on the bitonic sorting network and sketch a proof why they work.

\section{Differentiable Sorting Networks}

\newcommand{\steepness}{s}
\newcommand{\sigmoidfn}{\sigma}
\newcommand{\smoothTransSymbol}{\nu}

To relax sorting networks, we need to relax the $\min$ and $\max$ operators, which are used as a basis for the swap operations in sorting networks.
For that, we use $\operatorname{softmin}$ and $\operatorname{softmax}$, which are convex combinations via the logistic sigmoid function $\sigmoidfn(x)=\frac{1}{1+e^{-x}}\,$.
For two elements $a_i, a_{j}$, we define in accordance to $\operatorname{softmin}$ and $\operatorname{softmax}$:
\begin{align}
        \operatorname{softmin}(a_i, a_{j})\, &:=\ \qquad\  \alpha_{ij}\   \cdot a_i +  (1-\alpha_{ij}) \cdot a_{j}\\
        \operatorname{softmax}(a_i, a_{j})\, &:=\ (1-\alpha_{ij}) \cdot a_i +  \qquad \ \alpha_{ij}\  \cdot a_{j}
\end{align}
where
\begin{equation}
    \alpha_{ij} := \sigmoidfn((a_j - a_i) \cdot \steepness)
    \label{eq:exchange-alpha}
    .
\end{equation}
Here, $\steepness$ denotes a steepness hyperparameter such that for $\steepness\to\infty$ the smooth operators converge to the discrete operators.
As we show in the next section, it is necessary to extend this formulation by the activation replacement trick~$\varphi$ to avoid vanishing gradients and extensive blurring.

\begin{figure}[t]
    \centering
    \begin{tikzpicture}[node distance=50pt, p/.style={circle, minimum size=3pt, inner sep=0pt, outer sep=0pt, fill, anchor=center}]
        \colorlet{arrowColor}{black!65}
        \definecolor{originalDist}{RGB}{243, 155, 35}
        \definecolor{newDist}{RGB}{2, 115, 85}
        \definecolor{newDist2}{RGB}{115, 198, 54}
        \colorlet{newDist}{newDist!80!newDist2}
        \node[anchor=south] (sigmoid) at (0,0) {
            \begin{tikzpicture}[scale=1., every node/.style={scale=1.}]
                \draw[->] (-2.5,0) -- (2.7,0) node[right] {$x$};
                \draw[->] (0,0) -- (0,1.2) node[above] {$\sigma(x)$};
                \draw[scale=1.,domain=-2.5:2.5,smooth,variable=\x] plot ({\x},{1/(1+exp(-\x*1.25))});
                
                \foreach \x in {-2,...,2} {
                    \draw (\x,0) -- (\x,-.07) node[anchor=north] {$\x$};
                };
                
                \setcounter{arrown}{0}
                \foreach \x in {-2.5, -.5,.6,1.4} {
                    \node[p, originalDist] (begin\thearrown) at (\x, {1/(1+exp(-\x*1.25))}) {};
                    \node[p, newDist] (end\thearrown) at ({\x/abs(\x)^.6}, {1/(1+exp(-(\x/abs(\x)^.6)*1.25))}) {};
                    \stepcounter{arrown}
                };
                \draw[->,>=latex, arrowColor] (begin0) edge[bend left] (end0) ;
                \draw[->,>=latex, arrowColor] (begin1) edge[in=100, out=80, looseness=3.5] (end1) ;
                \draw[->,>=latex, arrowColor] (begin2) edge[in=100, out=135, looseness=4.5] (end2) ;
                \draw[->,>=latex, arrowColor] (begin3) edge[in=100, out=70, looseness=4] (end3) ;
            \end{tikzpicture}
        };
        \node[anchor=south] (distributions) at (0,-2.2) {
            \begin{tikzpicture}[scale=1., every node/.style={scale=1.}]
                \draw[->] (-3,0) -- (3.2,0) node[right] {$x$};
                \draw[->] (0,0) -- (0,1.2);
                \draw[scale=1.,domain=-3:3,smooth,variable=\x, originalDist] plot ({\x},{exp(-1/2*(\x/1.)^2)});
                \draw[scale=1.,domain=-3:3,smooth,variable=\x, newDist] plot ({\x},{exp(-1/2*(\x/1.)^2)*(2*abs(\x)^.5)*0.60814}); 
                
                \foreach \x in {-2,...,2} {
                    \draw (\x,0) -- (\x,-.07) node[anchor=north] {$\x$};
                };
                
                \setcounter{arrown}{0}
                \foreach \x in {-2.5, -.5,.6,1.4} {
                    \node[p, originalDist] (begin\thearrown) at ({\x},{exp(-1/2*(\x/1.)^2)}) {};
                    \node[p, newDist] (end\thearrown) at ({(\x/abs(\x)^.6)},{exp(-1/2*((\x/abs(\x)^.6)/1.)^2)*(2*abs((\x/abs(\x)^.6))^.5)*0.60814}) {};
                    \stepcounter{arrown}
                };
                \draw[->,>=latex, arrowColor] (begin0) edge[bend left] (end0) ;
                \draw[->,>=latex, arrowColor] (begin1) edge[in=100, out=80, looseness=3.5] (end1) ;
                \draw[->,>=latex, arrowColor] (begin2) edge[in=80, out=110, looseness=4.5] (end2) ;
                \draw[->,>=latex, arrowColor] (begin3) edge[in=65, out=55, looseness=3.5] (end3) ;
                
                \node[originalDist, anchor=east] at (3.5, .7) {$p(x)$};
                \node[newDist, anchor=east]      at (3.5, 1.25) {$p(\varphi (x))$};
            \end{tikzpicture}
        };
    \end{tikzpicture}
    \captionof{figure}{
    The Activation Replacement Trick. 
    Top: on the logistic sigmoid function, the input values $x$ (orange) are mapped to $\varphi(x)$ (green) and are thus closer to $-1$ and $+1$.
    Bottom: probability density functions of Gaussian distributed input values $x$ (orange) and the distribution of replaced input values $\varphi(x)$ (green).
    }
    \label{fig:act-replacement-trick}
\end{figure}
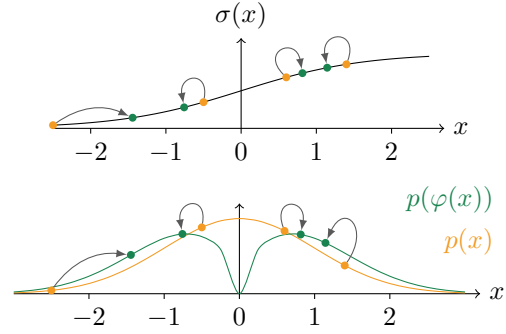

\subsection{Activation Replacement Trick \texorpdfstring{$\varphi$}{}}

Assuming that the inputs to a sorting network are normally distributed, there are many cases in which the differences of two values $|a_j -a_i|$ are very small as well as many cases in which the differences are very large.
For the relaxation of sorting networks, this poses two problems:\\[-1.3em]

If $|a_j -a_i|$ is close to $0$, while we obtain large gradients, this also blurs the two values to a great extent, modifying them considerably.
Thus, it is desirable to avoid $|a_{j} - a_i| \approx 0$.\\[-1.3em]

On the other hand, if $|a_{j} - a_i|$ is large, vanishing gradients occur, which hinders training.\\[-1.3em]

To counter these two problems at the same time, we propose the activation replacement trick.
We transform the differences between two values to be potentially swapped (e.g., $x=(a_j - a_i)$) from a unimodal Gaussian distribution into a bimodal distribution, which has a low probability density around $0$.
To this end, we apply the transformation\\[-.4em]
\newcommand{\ARTstrength}{\lambda}
\begin{equation}
    \varphi : x\mapsto \frac{x}{{|x|}^\ARTstrength + \epsilon}
    \label{eq:varphi-def}
\end{equation}\\[-.4em]
to the differences $x$, where $\ARTstrength \in [0, 1]$ and $\epsilon\approx 10^{-10}$. 
$\varphi$ pushes all input values (depending on the sign) toward $-1$ and $+1$, respectively.
Thus, by applying $\varphi$ before $\sigma$, we move the input values outside $[-1,+1]$ to positions at which they have a larger gradient, thus mitigating the problem of vanishing gradients. 
Simultaneously, we achieve a probability density of $0$ at $|a_j - a_i| = 0$ (i.e., here $p(\varphi(0)) = 0$) as all values close to zero are mapped toward $-1$ and $+1$, respectively.
This is displayed in Figure~\ref{fig:act-replacement-trick}.

As we multiply by the steepness parameter~$\steepness$ (Equation~\ref{eq:exchange-alpha}), we map the input to the sigmoid function toward $-\steepness$ and $+\steepness$, respectively.
Thus, when replacing $\sigma(x \cdot \steepness)$ by $\sigma(\varphi(x) \cdot \steepness)$, we push the output values toward $\frac{1}{1+e^{-1 \cdot s}}$ or $\frac{1}{1+e^{1 \cdot s}}$.
This increases the gradient $\frac{\partial \sigmoid(\varphi(x))}{\partial x}$ for large $\operatorname{abs}(x)$ which are those values causing the vanishing gradients, addressing the problem of vanishing gradients. 
Further, for all $x\in (-1, +1)$ this pushes the output values away from $\nicefrac{1}{2}$, addressing the problem of blurring of values. 

Therefore, we extend our formulation of the relaxations of the $\min$ and $\max$ operators by defining
\begin{equation}
    \alpha_{ij} := \sigmoidfn(\varphi(a_j - a_i) \cdot \steepness)
    .
\end{equation}
Empirically, the activation replacement trick accelerates the training through our sorting network.
We observe that, while sorting networks up to $21$~layers (i.e., bitonic networks with $n\leq 64$) can operate with moderate steepness (i.e., $\steepness \leq 15$) and without the activation replacement trick (i.e., $\ARTstrength=0$), for more layers, the activation replacement trick becomes necessary for good performance.
Notably, the activation replacement trick also improves the performance for sorting networks with fewer layers.
Further, the activation replacement trick allows training with smaller steepness $\steepness$, which makes training more stable specifically for long sequences as it avoids exploding gradients.

Note that, in case of bitonic, in the first layer of the last merge block, $\nicefrac{n}{2}$ elements in non-descending order are element-wise compared to $\nicefrac{n}{2}$ elements in non-ascending order. 
Thus, in this layer, we compare the minimum of the first sequence to the maximum of the second sequence and vice versa.
At the same time, we also compare the median of both sequences as well as values close to the median to each other.
While we consider very large differences as well as very small differences in the same layer, the activation replacement trick achieves an equalization of the mixing behavior, reducing blurring and vanishing gradients.

\subsection{Differentiable Permutation Matrices}

For sorting and ranking supervision, i.e., training a neural network to predict scalars, where only the order of these scalars is known, we use the ground truth permutation matrix as supervision.
Thus, to train an underlying neural network end-to-end through the differentiable sorting network, we need to return the underlying permutation matrix rather than the actual sorted scalar values.
For that, we compute the permutation matrices for the swap operations for each layer as shown in Figure~\ref{fig:overall-architecture-odd-even}.
Here, for all swap operations between any elements $a_i$ and $a_j$ that are to be ordered in non-descending order, the layer-wise permutation matrix is
\begin{align}
    P_{l,ii} = P_{l,jj} &= \phantom{1-{}}\alpha_{ij} = \phantom{1-{}} \sigmoidfn(\varphi(a_j - a_i) \cdot \steepness), \\
    P_{l, ij}= P_{l,ji} &= 1-\alpha_{ij} = 1 - \sigmoidfn(\varphi(a_j - a_i) \cdot \steepness)\,
\end{align}
where all other entries of $P_l$ are set to $0$.
By multiplication, we compute the complete relaxed permutation matrix $\pmb{P}$ as
\begin{equation}
    \pmb{P} = P_{n} \cdot ...\cdot P_2\cdot P_1 = \bigg(\prod_{l=1}^{n} P_l^\top\bigg)^{\!\top}\,.
    \label{eq:p-mult}
\end{equation}
A column in the relaxed permutation matrix can be seen as a distribution over possible ranks for the corresponding input value. 
Given a ground truth permutation matrix $\pmb{Q}$, we can define our column-wise cross entropy loss as
\begin{equation}
    \mathcal{L} := \sum_{c=1}^n \left( \frac{1}{n} \operatorname{CE}\left(\pmb{P}_c, \pmb{Q}_c\right) \right)
    \label{eq:loss}
\end{equation}
where $\pmb{P}_c$ and $\pmb{Q}_c$ denote the $c$th columns of $\pmb{P}$ and $\pmb{Q}$, respectively.
Note that, as the cross entropy loss is, by definition, computed element-wise, the column-wise cross entropy is equivalent to the row-wise cross entropy.

\begin{table*}[t]
    \caption{
    Results for the comparison to state-of-the-art \cite{Grover2019-NeuralSort, Cuturi2019-SortingOT} using the same network architectures averaged over 5 runs.
    The first three rows are duplicated from \citet{Cuturi2019-SortingOT}.
    Metrics are \hbox{(EM | EW | EM5)}.
    }
    \label{table:results-mnist}
    \centering
    \newcommand{\emfivespacer}[0]{\phantom{\ 00.0}}
    \small
    \begin{tabular}{lccccc}
        \toprule
        Method & $n=\pmb{3}$ & $n=\pmb{5}$ & $n=\pmb{7}$ & $n=\pmb{9}$ & $n=\pmb{15}$ \\
        \midrule
        Stoch.~NeuralSort       & $92.0\ |\ 94.6\ |\emfivespacer{}$ & ${79.0}\ |\ 90.7\ |\ {79.0}$ & $63.6\ |\ 87.3\ |\emfivespacer{}$ & $45.2\ |\ 82.9\ |\emfivespacer{}$ & $12.2\ |\ 73.4\ |\emfivespacer{}$  \\
        Det.~NeuralSort         & $91.9\ |\ 94.5\ |\emfivespacer{}$ & ${77.7}\ |\ 90.1\ |\ {77.7}$ & $61.0\ |\ 86.2\ |\emfivespacer{}$ & $43.4\ |\ 82.4\ |\emfivespacer{}$ & $\phantom{0}9.7\ |\ 71.6\ |\emfivespacer{}$  \\
        Optimal Transport       & $92.8\ |\ 95.0\ |\emfivespacer{}$ & ${81.1}\ |\ 91.7\ |\ {81.1}$ & $65.6\ |\ 88.2\ |\emfivespacer{}$ & $49.7\ |\ 84.7\ |\emfivespacer{}$ & $12.6\ |\ 74.2\ |\emfivespacer{}$  \\
        \midrule
        Fast Sort \& Rank       & $90.6\ |\ 93.5\ |\ 73.5$  &  $71.5\ |\ 87.2\ |\ 71.5$  &  $49.7\ |\ 81.3\ |\ 70.5$  &  $29.0\ |\ 75.2\ |\ 69.2$  &  $\phantom{0}2.8\ |\ 60.9\ |\ 67.4$ \\
        \midrule
            Odd-Even      & $\pmb{95.2}\ |\ \pmb{96.7}\ |\ \pmb{86.1}$ & $\pmb{86.3}\ |\ \pmb{93.8}\ |\ \pmb{86.3}$ & $\pmb{75.4}\ |\ \pmb{91.2}\ |\ \pmb{86.4}$ & $\pmb{64.3}\ |\ \pmb{89.0}\ |\ \pmb{86.7}$ & $\pmb{35.4}\ |\ \pmb{83.7}\ |\ \pmb{87.6}$  \\
        \bottomrule
        \toprule
         & $n=\pmb{2}$ & $n=\pmb{4}$ & $n=\pmb{8}$ & $n=\pmb{16}$ & $n=\pmb{32}$  \\
        \midrule
            Odd-Even        & $98.1\ |\ 98.1\ |\ 84.3$ & $90.5\ |\ 94.9\ |\ 85.5$ & $63.6\ |\ 87.9\ |\ 83.6$ & $31.7\ |\ 82.8\ |\ 87.3$ & $\phantom{0}1.7\ |\ 69.1\ |\ 86.7$  \\
            Bitonic      & $98.1\ |\ 98.1\ |\ {84.0}$ & $91.4\ |\ 95.3\ |\ 86.7$ & $70.6\ |\ 90.3\ |\ {86.9}$ & $30.5\ |\ 81.7\ |\ 86.6$ & $\phantom{0}2.7\ |\ 67.3\ |\ 85.4$  \\
        \bottomrule
    \end{tabular}%
\end{table*}

\begin{table*}[t]
    \caption{Results for training on the SVHN data set averaged over 5 runs. Metrics are \hbox{(EM | EW | EM5)}.}
    \label{tab:svhn}
    \centering
    \small
    \newcommand{\emfivespacer}[0]{\phantom{\ 00.0}}
    \newcommand{\minispace}[0]{}
    \begin{tabular}{lccccc}
        \toprule
        $\quad~~~~$Method$\quad~~~$ & $n=\pmb{2}$ & $n=\pmb{4}$ & $n=\pmb{8}$ & $n=\pmb{16}$ & $n=\pmb{32}$  \\
        \midrule
        Det.~NeuralSort     & $90.1\ |\ 90.1\ |\ 39.9$  &  $61.4\ |\ 78.1\ |\ 45.4$  &  $15.7\ |\ 62.3\ |\ 48.5$  &  $\phantom{0}0.1\ |\ 45.7\ |\ 51.0$  &  $\phantom{0}0.0\ |\ 29.9\ |\ 52.7$  \\
        Optimal Transport   & $85.5\ |\ 85.5\ |\ 25.9$  &  $57.6\ |\ 75.6\ |\ 41.6$  &  $19.9\ |\ 64.5\ |\ 51.7$  &  $\phantom{0}0.3\ |\ 47.7\ |\ 53.8$  &  $\phantom{0}0.0\ |\ 29.4\ |\ 53.3$  \\
        Fast Sort \& Rank   & $93.4\ |\ 93.4\ |\ 57.6$  &  $58.0\ |\ 75.8\ |\ 41.5$  &  $\phantom{0}8.6\ |\ 52.7\ |\ 34.4$  &  $\phantom{0}0.3\ |\ 36.5\ |\ 41.6$  &  $\phantom{0}0.0\ |\ 14.0\ |\ 27.5$  \\
        \midrule
            Odd-Even      & \minispace{}$93.4\ |\ 93.4\ |\ 58.0$\minispace{} & \minispace{}$\pmb{74.8}\ |\ \pmb{85.5}\ |\ \pmb{62.6}$\minispace{} & \minispace{}$35.2\ |\ 73.5\ |\ 63.9$\minispace{} & \minispace{}$\phantom{0}1.8\ |\ 54.4\ |\ 62.3$\minispace{} & \minispace{}$\phantom{0}0.0\ |\ 36.6\ |\ 62.6$\minispace{}  \\
            Bitonic       & $\pmb{93.8}\ |\ \pmb{93.8}\ |\ \pmb{58.6}$ & $74.4\ |\ 85.3\ |\ 62.1$ & $\pmb{38.3}\ |\ \pmb{75.1}\ |\ \pmb{66.8}$ & $\pmb{\phantom{0}3.9}\ |\ \pmb{59.6}\ |\ \pmb{66.8}$ & $\phantom{0}0.0\ |\ \pmb{42.4}\ |\ \pmb{67.7}$  \\
        \bottomrule
    \end{tabular}%
\end{table*}

\section{Experiments\protect\footnote{Our implementation is openly available at \href{https://github.com/Felix-Petersen/diffsort}{github.com/Felix-Petersen/diffsort}.}}

\subsection{Sorting and Ranking Supervision}

We evaluate the proposed differentiable sorting networks on the four-digit MNIST sorting benchmark \cite{Grover2019-NeuralSort,Cuturi2019-SortingOT} as well as on the real-world SVHN data set. %

\makeatletter
\def\subsubsection{\@startsection{subsubsection}{3}{\z@}{-0.08in}{0.01in}
                {\normalsize\bf\raggedright}}
\makeatother

\paragraph{MNIST} 
For the four-digit MNIST sorting benchmark, MNIST digits are concatenated to four-digit numbers, e.g., \mnistimg{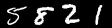}. 
A CNN then predicts a scalar value corresponding to the value displayed in the four-digit image.
For training, $n$ of those four-digit images are separately processed by the CNN and then sorted by the relaxed sorting network as shown in Figure~\ref{fig:overall-architecture-odd-even}.
Based on the permutation matrix produced by the sorting network and the ground truth ranking, the training objective is computed (Equation~\ref{eq:loss}) and the CNN is updated.
At test time, we forward single images of four-digit numbers from the test data set. 
For evaluation, the discrete rankings of the predicted values are compared to the rankings of their ground truth.
Note that the $n$ used for testing and evaluation can be independent of the $n$ used for training because the $n$ images are processed independently.

\paragraph{SVHN}
Since the multi-digit MNIST data set is an artificial data set, we also evaluate our technique on the SVHN data set \cite{Netzer2011-SVHN}. 
This data set comprises house numbers collected from Google Street View and provides a larger variety wrt.~different fonts and formats than the MNIST data set.
We use the published ``Format~$1$'' and preprocess it as described by \citet{Goodfellow2013-Multi-digit-SVHN}, cropping the centered multi-digit numbers with a boundary of $30\%$, resizing it to a resolution of $64\times64$, and then selecting $54\times54$ pixels at a random location.
As SVHN contains $1-5$ digit numbers, we can avoid the concatenation and use the original images directly.
Example images are
~\svhnimg{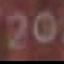}
~\svhnimg{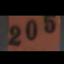}
~\svhnimg{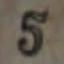}
~\svhnimg{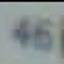}
~\svhnimg{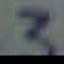}
~\svhnimg{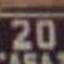}
~\svhnimg{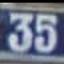}
~\svhnimg{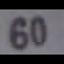}
~\svhnimg{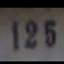}
~\svhnimg{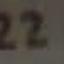}%
\;.
Otherwise, the experimental setup is as for the four-digit MNIST data set.

\paragraph{Network Architecture} For the MNIST sorting task,  we use the same convolutional neural network (CNN) architecture as \citet{Grover2019-NeuralSort} and \citet{Cuturi2019-SortingOT} to allow for comparability.
This architecture consists of two convolutional layers with a kernel size of $5\times5$, $32$ and $64$ channels respectively, each followed by a ReLU and MaxPool layer; 
this is (after flattening) followed by a fully connected layer with a size of $64$, a ReLU layer, and a fully connected output layer mapping to a scalar.

For the SVHN task, we use a network with four convolutional layers with a kernel size of $5\times5$ and ($32, 64, 128, 256$) filters, each followed by a ReLU and a max-pooling layer with stride $2\times2$; followed by a fully connected layer with size $64$, a ReLU, and a layer with output size $1$.

\paragraph{Evaluation Metrics} 
For evaluation, discrete rankings based on the scalar predictions are computed and compared to the discrete ground truth rankings.
As in previous works, we use the evaluation metrics of exact match (EM) of the predicted ranking, and fraction of element-wise correct ranks (EW) in the predicted ranking.  
For EM and EW, we follow \citet{Grover2019-NeuralSort} and \citet{Cuturi2019-SortingOT}, and use the same $n$ for training and evaluation.
However, this can be a problem in the context of large input sets as these evaluation metrics become unreliable as $n$ increases.
For example, the difficulty of exact matches rises with the factorial of $n$, which is why they become too sparse to allow for valid conclusions for large $n$. 
To allow for a comparison of the performance independent of the number of elements~$n$ used for training, we also evaluate the models based on the EM accuracy for $n=5$ (EM5). 
That is, the network can be trained with an arbitrary $n$, but the evaluation is done for $n=5$.
A table with respective standard deviations can be found in Supplementary Material \ref{sm:std}.

\paragraph{Training Settings} 
We use the Adam optimizer~\cite{Kingma2014AdamOpt} with a learning rate of $10^{-3.5}$, and up to $10^6$ steps of training.
Furthermore, we set $\ARTstrength=0.25$ and use a steepness of two times the number of layers ($\steepness=2n$ for odd-even and $\steepness=(\log_2n)(1+\log_2n)$ for bitonic.)
We use a constant batch size of $100$ as in previous works unless denoted otherwise.
Note that, although $\ARTstrength$ is chosen as a constant value for all $n$, a higher accuracy is possible when optimizing $\ARTstrength$ for each $n$ separately.

\begin{table*}[t]
    \centering
    \caption{
    Results for large $n$ measured using the EM5 metric with fixed number of samples as well as a fixed batch size.
    Independent of the batch size, the model always performs better for larger $n$.
    Trained for $10^4$ steps \& averaged over $10$ runs.
    }
    \label{table:results-bitonic-large}
    \footnotesize
    \resizebox{1.\textwidth}{!}{
    \begin{tabular}{l|c|cccccc|cccccc}
        \toprule
        $\qquad\ARTstrength$      & $0.25$    & $0.4$     & $0.4$     & $0.4$     & $0.4$     & $0.4$     & $0.4$     & $0.4$     & $0.4$     & $0.4$     & $0.4$     & $0.4$     & $0.4$  \\
        \midrule
        $\qquad n$          & $32$      & $32$      & $64$      & $128$     & $256$     & $512$     & $1024$    & $32$      & $64$      & $128$     & $256$     & $512$     & $1024$  \\
        \midrule
        batch size          & $128$     & $128$     & $64$      & $32$      & $16$      & $8$       & $4$       & $4$       & $4$       & $4$       & $4$       & $4$       & $4$  \\
        \midrule
            $\steepness=30$   & $\pmb{78.20}$ & $79.89$ & $81.25$ & $\pmb{82.50}$ & $\pmb{82.05}$ & $82.50$ & $\pmb{82.80}$ & $71.08$ & $\pmb{75.88}$ & $79.43$ & $\pmb{81.46}$ & $82.98$ & $\pmb{82.80}$  \\
            $\steepness=32.5$ & $76.98$ & $79.62$ & $\pmb{81.66}$ & $80.15$ & $81.87$ & $82.64$ & $81.63$ & $\pmb{72.31}$ & $75.59$ & $\pmb{79.71}$ & $81.36$ & $\pmb{82.99}$ & $81.63$  \\
            $\steepness=35$   & $77.45$ & $80.93$ & $81.26$ & $80.72$ & $81.42$ & $81.51$ & $81.15$ & $71.15$ & $75.73$ & $78.81$ & $79.32$ & $82.30$ & $81.15$  \\
            $\steepness=37.5$ & $76.40$ & $80.02$ & $80.05$ & $81.50$ & $80.05$ & $\pmb{82.67}$ & $80.07$ & $70.69$ & $75.80$ & $79.11$ & $80.64$ & $82.70$ & $80.07$  \\
            $\steepness=40$   & $77.69$ & $\pmb{80.97}$ & $80.23$ & $81.55$ & $79.75$ & $81.89$ & $81.15$ & $70.20$ & $74.67$ & $78.14$ & $80.06$ & $81.39$ & $81.15$  \\
            \midrule
            mean              & $77.35$ & $80.29$ & $80.89$ & $81.28$ & $81.03$ & $82.24$ & $81.36$ & $71.09$ & $75.53$ & $79.04$ & $80.57$ & $82.47$ & $81.36$  \\
            best $s$          & $\pmb{78.20}$ & $\pmb{80.97}$ & $\pmb{81.66}$ & $\pmb{82.50}$ & $\pmb{82.05}$ & $\pmb{82.67}$ & $\pmb{82.80}$ & $\pmb{72.31}$ & $\pmb{75.88}$ & $\pmb{79.71}$ & $\pmb{81.46}$ & $\pmb{82.99}$ & $\pmb{82.80}$  \\
            worst $s$         & $76.40$ & $79.62$ & $80.05$ & $80.15$ & $79.75$ & $81.51$ & $80.07$ & $70.20$ & $74.67$ & $78.14$ & $79.32$ & $81.39$ & $80.07$  \\
        \bottomrule
    \end{tabular}%
    }
\end{table*}

\subsubsection{Results}

\paragraph{Comparison to State-of-the-Art (MNIST)}
We first compare our approach to the methods proposed by \citet{Grover2019-NeuralSort} and \citet{Cuturi2019-SortingOT}.
Here, we follow the setting that the $n$ used for evaluation is the same as the $n$ used for training. 
The evaluation is shown in Table~\ref{table:results-mnist}. 
We report results for exact match, correct ranks, and EM5, respectively. 
For the odd-even architecture, we compare results for the original $n\in\{3,5,7,9,15\}$. 
Our approach outperforms current methods on all metrics and input set sizes. 
In addition, we extend the original benchmark set sizes by $n\in\{2,4,8,16,32\}$, allowing for the canonical version of the bitonic sorting network which requires input size of powers of $2$.
We apply $n\in\{2,4,8,16,32\}$ to the odd-even as well as the bitonic sorting network.
In this direct comparison, we can see that the bitonic and the odd-even architectures perform similar.
Notably, the EM and EW accuracies do not always correlate as can be seen for $n=32$. 
Here, the EM accuracy is greater for the bitonic network and the EW accuracy is greater for the odd-even network.
We attribute this to the odd-even network's gradients causing swaps of neighbors while the bitonic network's gradients provide a holistic approach favoring exact matches. 

\label{sec:evaluation_svhn}
\paragraph{SVHN}
The results in Table~\ref{tab:svhn} show that the real-world SVHN task is significantly harder than the MNIST task.
On this data set, differentiable sorting networks are also better than current methods on all metrics and input set sizes. 
Here, the performances of odd-even and bitonic are similar.
Notably, the EM5 accuracy is largest for the bitonic sorting network at $n=32$, which demonstrates that the method benefits from longer input sets.
Further, for $n\in\{8, 16, 32\}$, the bitonic sorting network marginally outperforms the odd-even sorting network on all metrics.

\subsection{Large-Scale Sorting and Ranking Supervision}

We are interested in the effect of training with larger input set sizes $n$. 
As the bitonic sorting network requires significantly fewer layers than odd-even and is (thus) faster, we use the bitonic sorting network for the scalability experiments.
Here, we evaluate for $n=2^k, k\in\{5,6,7,8,9,10\}$ on the MNIST sorting benchmark, comparing the EM5 accuracy as shown in Table \ref{table:results-bitonic-large}.

For this experiment, we consider steepness values of $\steepness\in\{30, 32.5, 35, 37.5, 40\}$ and report the mean, best, and worst over all steepness values for each $n$.
We set $\ARTstrength$ to $0.4$ as this allows for stable training with $n>128$. 
To keep the evaluation feasible, we reduce the number of steps during training to $10^4$, compared to the $10^6$ iteration in Table~\ref{table:results-mnist}. 
Again, we use the Adam optimizer with a learning rate of $10^{-3.5}$.

In the first two columns of Table~\ref{table:results-bitonic-large}, we show a head-to-head comparison with the setting in Table~\ref{table:results-mnist} with $\ARTstrength=0.25$ and $\ARTstrength=0.4$ for $n=32$.
Trained for $10^6$ steps, the EM5 accuracy is $85.4\%$, while it is $78.2\%$ after $10^4$ steps.
Increasing $\ARTstrength$ from $0.25$ to $0.4$ improves the EM5 accuracy from $78.2\%$ to $80.97\%$.

This also demonstrates that already at this scale, a larger $\ARTstrength$, i.e., a stronger activation replacement trick, can improve the overall accuracy of a bitonic sorting network compared to training with $\ARTstrength=0.25$.

As the size of training tuples $n$ increases, this also increases the overall number of observed images during training.
Therefore, in the left half of Table~\ref{table:results-bitonic-large}, we consider the accuracy for a constant total of observed images per iteration, i.e., for $n\times\mathrm{batch~size} = 4096$ (e.g., for $n=32$ this results in a batch size of $128$, while for $n=1024$, the batch size is only $4$). 
In the right half of Table~\ref{table:results-bitonic-large}, we consider a constant batch size of $4$.

With increasing $n$, the accuracy of our model increases even for a constant number of observed images even though it has to operate on very small batch sizes. 
This shows that training with larger ordered sets results in better accuracy.
This suggests that, if possible, larger $n$ should be prioritized over larger batch sizes and that good results can be achieved by using the largest possible $n$ for the available data to learn from all available information.

\begin{table*}[t]
    \caption{
    Runtimes, memory requirements, and number of layers for sorting $n$ elements. %
    Runtimes reported for an Nvidia GTX 1070. We include NeuralSort \cite{Grover2019-NeuralSort}, FastRank \cite{Blondel2020-FastSorting}, and OT Sort \cite{Cuturi2019-SortingOT}.
    }
    \label{tab:runtime-memory}
    \centering
    \newcommand{\mBi}{{\footnotesize$\mathrm{B}$}}
    \newcommand{\mKB}{{\footnotesize$\mathrm{KB}$}}
    \newcommand{\mMB}{{\footnotesize$\mathrm{MB}$}}
    \newcommand{\mGB}{{\footnotesize$\mathrm{GB}$}}
    \newcommand{\rtbold}[1]{{\bm{#1}}}
    \resizebox{1.\linewidth}{!}{
    \begin{tabular}{crrrrrrrrrrrrrrrrrrr}
        \toprule 
                & \multicolumn{4}{c}{Differentiable Odd-Even Sort}                    & \multicolumn{4}{c}{Differentiable Bitonic Sort}                    && \multicolumn{2}{c}{NeuralSort}                    & {FastRank}                    & {OT Sort}                          \\
            \cmidrule(r){2-5}\cmidrule(r){6-9}\cmidrule(r){11-12}\cmidrule(r){13-13}\cmidrule(r){14-14}
        $n$     & GPU & CPU & Memory & \# Layers                                          & GPU & CPU & Memory & \# Layers                                                  && GPU & CPU                                         & CPU                           & CPU                                            \\
        \midrule 
        $4$      & $\rtbold{69\,\mathrm{ns}}$     &  $1.9\,\mu\mathrm{s}$    & $1$\mKB     & $4$        & $\rtbold{52\,\mathrm{ns}}$       & $1.3\,\mu\mathrm{s}$     & $840$\mBi  & $3$            && $\rtbold{145\,\mathrm{ns}}$       & $7.1\,\mu\mathrm{s}$  & $\rtbold{189\,\mu\mathrm{s}}$          & $\rtbold{1.0\,\mathrm{ms}}$   \\
        $16$     & $\rtbold{1.2\,\mu\mathrm{s}}$   &   $54\,\mu\mathrm{s}$   & $42$\mKB    & $16$       & $\rtbold{759\,\mathrm{ns}}$      & $40\,\mu\mathrm{s}$      & $28$\mKB  & $10$            && $\rtbold{396\,\mathrm{ns}}$       & $11\,\mu\mathrm{s}$   & $\rtbold{215\,\mu\mathrm{s}}$          & $\rtbold{7.5\,\mathrm{ms}}$   \\
        $32$     & $\rtbold{7.4\,\mu\mathrm{s}}$  &  $309\,\mu\mathrm{s}$    & $315$\mKB   & $32$       & $\rtbold{3.5\,\mu\mathrm{s}}$    & $159\,\mu\mathrm{s}$     & $152$\mKB & $15$            && $\rtbold{969\,\mathrm{ns}}$       & $13\,\mu\mathrm{s}$   & $\rtbold{303\,\mu\mathrm{s}}$          & $\rtbold{17\,\mathrm{ms}}$      \\
        $128$    & $\rtbold{493\,\mu\mathrm{s}}$  & $19\,\mathrm{ms}$        & $20.2$\mMB  & $128$      & $\rtbold{97\,\mu\mathrm{s}}$     & $5\,\mathrm{ms}$         & $4.1$\mMB & $28$            && $\rtbold{12\,\mu\mathrm{s}}$      & $177\,\mu\mathrm{s}$  & $\rtbold{834\,\mu\mathrm{s}}$          & $\rtbold{55\,\mathrm{ms}}$        \\
        $1\,024$ & $\rtbold{660\,\mathrm{ms}}$     & $31\,\mathrm{s}$         & $4.9$\mGB  & $1\,024$   & $\rtbold{15\,\mathrm{ms}}$       & $1.7\,\mathrm{s}$        & $549$\mMB & $55$            && $\rtbold{1.2\,\mathrm{ms}}$       & $11\,\mathrm{ms}$     & $\rtbold{4.8\,\mathrm{ms}}$            & $\rtbold{754\,\mathrm{ms}}$      \\
        \bottomrule
    \end{tabular}%
    }
\end{table*}

\subsection{Ablation Study and Hyperparameter Sensitivity}

To assess the impact of the proposed activation replacement trick (ART), we evaluate both architectures with and without ART at $\ARTstrength=0.25$ in Table~\ref{tab:ablation-study}.
The accuracy improves by using the ART for small as well as for large $n$.
For large $n$, the activation replacement trick has a greater impact on the performance of both architectures.
In Figure~\ref{fig:different-steepnesses}, we evaluate the sensitivity of the differentiable odd-even sorting network to the steepness hyperparameter $s$.
For a broad range of $s$, the performance is stable.
In Figure~\ref{fig:different-art-strengths}, we evaluate both differentiable sorting networks for varying ART intensities~$\ARTstrength$. 
Here, performance increases with larger $\ARTstrength$s (i.e., with a stronger ART). For $\ARTstrength > 0.5$, the performance drops as $\varphi$ converges to a discrete step function for $\ARTstrength \to 1$.

\subsection{Top-$k$ Supervision}

In addition to the sorting supervision task, we also benchmark our method on top-$k$ supervision following \citet{Cuturi2019-SortingOT} and \citet{Blondel2020-FastSorting}.
Here, we train two models (ResNet18 and a vanilla CNN with 4 convolutional and 2 fully connected layers) on CIFAR-10 as well as CIFAR-100 and compare the results to training with the Softmax Cross-Entropy loss. 
Further details on the experimental setting can be found in Supplementary Material~\ref{sm:topk}.
Following \citet{Cuturi2019-SortingOT} and \citet{Blondel2020-FastSorting}, we focus on $k=1$.
We present the results for this in Table~\ref{tab:top-k}.
Overall, Softmax Cross-Entropy and our differentiable top-$k$ operator perform similar even in the $100$ class classification problem.

\subsection{Runtime and Memory Analysis}
\label{sec:evaluation-runtime}
Finally, we report the runtime and memory consumption of differentiable sorting networks in Table~\ref{tab:runtime-memory}. 
For GPU runtimes, we use a native CUDA implementation and measure the time and memory for sorting $n$ input elements including forward and backward pass. 
For CPU runtimes, we use a PyTorch \cite{2019-PyTorch} implementation. 
For a small number of input elements, the odd-even and bitonic sorting networks have around the same time and memory requirements, while for larger numbers of input elements, bitonic is much faster than odd-even.

\begin{figure}[H]
    \centering
    \captionof{table}{
    Ablation Study: Evaluation of the ART ($\ARTstrength=0$ vs.~$\ARTstrength=0.25$) for $n=4$ and $n=32$ on the MNIST and the SVHN data set.
    The displayed metric is EW.
    \label{tab:ablation-study}
    }
    \vspace*{-.5em}
    {\small
    \begin{tabular}{lcccc}
        \toprule
        & \multicolumn{2}{c}{$n = 4$} & \multicolumn{2}{c}{$n = 32$} \\
        \cmidrule(l){2-3}\cmidrule(l){4-5}
        Setting~~/~~$\ARTstrength$ & $0$ & $0.25$ & $0$ & $0.25$ \\
        \midrule
        Odd-Even (MNIST)    & $94.5$ & $\pmb{94.9}$ & $61.5$ & $\pmb{69.1}$  \\
        Bitonic (MNIST)     & $93.6$ & $\pmb{95.3}$ & $62.8$ & $\pmb{67.3}$  \\
        \midrule
        Odd-Even (SVHN)     & $77.3$ & $\pmb{85.5}$ & $28.5$ & $\pmb{36.6}$  \\
        Bitonic (SVHN)      & $78.1$ & $\pmb{85.3}$ & $35.0$ & $\pmb{42.4}$  \\
        \bottomrule
    \end{tabular}}\\[1.em]
    \captionof{figure}{
    Sensitivity of the odd-even sorting network to varying steepness $s$ for $n=16$.
    \label{fig:different-steepnesses}
    }\vspace*{-.5em}
    \resizebox{\linewidth}{!}{
    \input{fig/plot_rebuttal_1.pgf}
    }
    \captionof{figure}{
    Comparing different ART strengths $\ARTstrength$ for $n=8$ (top) and $n=16$ (bottom). Training with $\lambda \leq 0.5$ is stable.%
    \label{fig:different-art-strengths}%
    }\vspace*{-.5em}
    \resizebox{\linewidth}{!}{
    \input{fig/plot_rebuttal_2.pgf}
    }
    \captionof{table}{
    Top-$k$ classification averaged over 10 runs.
    \label{tab:top-k}
    }
    \centering
    {\small
    \begin{tabular}{lcccc}
        \toprule
        Setting                         & Softmax CE & Diff.~Top-$k$ \\
        \midrule
        CIFAR-10, Vanilla CNN           & $87.2\%$   & $\pmb{88.0\%}$               \\
        CIFAR-10, ResNet18              & $\pmb{91.0\%}$   & $90.9\%$                 \\
        CIFAR-100, Vanilla CNN          & $\pmb{58.2\%}$   & $56.3\%$                 \\
        CIFAR-100, ResNet18             & $61.9\%$   & $\pmb{63.3\%}$                   \\
        \bottomrule
    \end{tabular}}
\end{figure}

The asymptotic runtime of differentiable odd-even sort is in $\mathcal{O}(n^3)$ and for bitonic sort the runtime is in $\mathcal{O}(n^2 (\log n)^2)$. 
Note that, for this, the matrix multiplication in Equation~\ref{eq:p-mult} is a sparse matrix multiplication.
We also report runtimes for other differentiable sorting and ranking methods.
For large $n$, we empirically confirm that FastRank \cite{Blondel2020-FastSorting} is the fastest method, i.a., because it produces only output ranks / sorted output values and not differentiable permutation matrices.
Note that differentiable sorting networks also produce sorted output values. 
Computing only sorted output values is significantly faster than computing the full differentiable permutation matrices, however, for the effective cross-entropy training objective, differentiable permutation matrices are necessary.

\section{Conclusion}

In this work, we presented differentiable sorting networks for training based on sorting and ranking supervision. 
To this end, we approximated the discrete $\min$ and $\max$ operators necessary for pairwise swapping in traditional sorting network architectures with their respective differentiable $\operatorname{softmin}$ and $\operatorname{softmax}$ operators.
We proposed an activation replacement trick to avoid the problems of vanishing gradients and well as blurred values. 
We showed that it is possible to robustly sort and rank even long sequences on large input sets of up to at least $1024$ elements.
In the future, we will investigate differentiable sorting networks for applications such as clustering and learning-to-rank.

\section*{Acknowledgment}

The second author gratefully acknowledges the financial support from
Land Salzburg within the WISS 2025 project IDA-Lab (20102-F1901166-KZP
and 20204-WISS/225/197-2019).

\bibliography{manual}
\bibliographystyle{icml2021}
\newpage~ %

\newif\ifappendix
\appendixtrue
\ifappendix

\renewcommand{\textfraction}{0.0}

\appendix

\begin{figure*}[t]
\newdimen\myunit\myunit1.15mm
\centering
\includegraphics[scale=1.15]{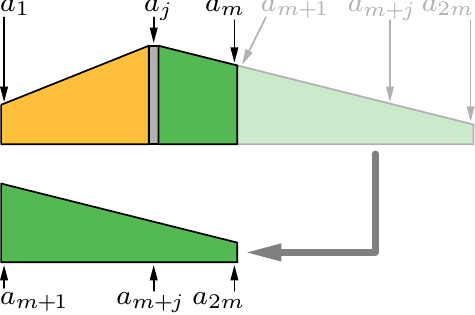}\hfill
\includegraphics[scale=1.15]{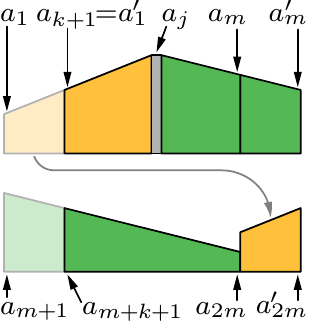}\hfill
\includegraphics[scale=1.15]{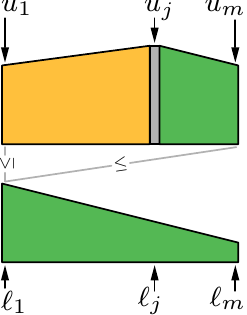}\hskip4mm
\includegraphics[scale=1.15]{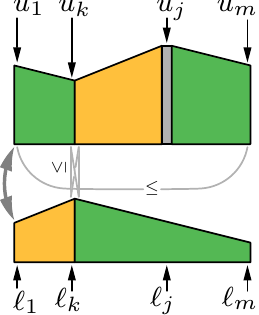} \\
\hbox to\textwidth{\footnotesize
\hbox to48.23\myunit{\hss (a)\hss}\hfill
\hbox to31.07\myunit{\hss (b)\hss}\hfill
\hbox to25.01\myunit{\hss (c)\hss}\hskip4mm
\hbox to24.91\myunit{\hss (d)\hss}}
\caption{Bitonic merge turns a bitonic input sequence into two
         bitonic output sequences, with all elements in the
         one (upper, $u_i$) sequence larger than all elements
         in the other (lower, $\ell_i$) sequence.
         The diagrams show the vertical alignment of elements to compare (a) and the 
         invariance to cyclic permutations (b).
         Depending on the values in (a), no exchanges (c) or exchanges (d) are executed.%
         \label{fig:bitonic_merge}}
\end{figure*}

\newpage

\section{The Bitonic Sorting Network}

\label{sm:bitonic}

In the following, we detail the bitonic sorting network and sketch a proof of why the bitonic sorting networks sorts:

A bitonic sequence is sorted by several bitonic merge blocks, shown in orange and green in Figure~\ref{fig:bitonic_sort}.
Each block takes a bitonic input sequence $a_1a_2\ldots a_{2m}$ of length~$2m$ and
turns it into two bitonic output sequences $\ell_1\ell_2\ldots\ell_m$ and $u_1u_2\ldots u_m$ of length~$m$ that satisfy
$\max_{i=1}^m l_i \le \min_{i=1}^m u_i$. 
These subsequences are recursively processed by bitonic merge blocks, until the output sequences are of length~$1$.
At this point, the initial bitonic sequence has been turned into a monotonic sequence due to the minimum/maximum conditions that hold between the output sequences (and thus elements).

A bitonic merge block computes its output as
$\ell_i = \min(a_i,a_{m+i})$ and $u_i = \max(a_i,a_{m+i})$.
This is depicted in Figure~\ref{fig:bitonic-schema} by the arrows pointing from the minimum to the maximum.
To demonstrate that bitonic merge works, we show that this
operation indeed produces two bitonic output sequences for which the relationship
$\max_{i=1}^m l_i \le \min_{i=1}^m u_i$ holds. 

Note that neither a cyclic permutation of the sequence ($a_i' = a_{(i+k-1 \bmod 2m)+1}$ for some $k$, Figure~\ref{fig:bitonic_merge}b), nor
a reversal, change the bitonic character of the sequence.
As can be seen in Figure~\ref{fig:bitonic_merge}b, even under cyclic permutation still the same pairs of elements are considered for a potential swap. 
Thus, as a cyclic permutation or a reversal only causes the output sequences to be analogously cyclically permuted or reversed, this changes neither the bitonic character of these sequences nor the relationship between them. 
Therefore, it suffices to consider the special case shown
in Figure~\ref{fig:bitonic_merge}a, with a monotonically increasing sequence (orange)
followed by a monotonically decreasing sequence (green) and the maximum
element~$a_j$ (gray) in the first half. Note that in this case
$\forall i; j \le i \le m: a_i \ge a_{m+i}
\wedge u_i = a_i \wedge \ell_i = a_{m+i}$.

For this case, we have to distinguish two sub-cases: \\
$a_1 \ge a_{m+1}$ and $a_1 < a_{m+1}$.

If, on one hand, $a_1 \ge a_{m+1}$, we have the situation shown in Figure \ref{fig:bitonic_merge}c: the output sequence $u_1u_2\ldots u_m$ is simply the first
half of the sequence, the output sequence $\ell_1\ell_2\ldots\ell_m$
is the second half. 
Thus, both output sequences are bitonic (since
they are subsequences of a bitonic input sequence) and
$\min_{i=1}^m u_i = \min(u_1,u_m) \ge \ell_1 = \max_{i=1}^m \ell_i$.

If, on the other hand, $a_1 < a_{m+1}$, we can infer
$\exists k; 1 \le k < j: a_k > a_{m+k} \wedge a_{k+1} \le a_{m+k+1}$.
This situation is depicted in Figure~\ref{fig:bitonic_merge}d. 
Thus,
$\forall i; 1 \le i \le k: u_i = a_{m+i} \wedge \ell_i = a_i$ and
$\forall i; k {\,<\,} i {\,\le\,} m: u_i {\,=\,} a_i \wedge \ell_i {\,=\,} a_{m+i}$. Since
$u_k = a_{m+k} > a_k = \ell_k$,
$u_k = a_{m+k} \ge a_{m+k+1} = \ell_{k+1}$,
$u_{k+1} \kern-0.8pt=\kern-0.8pt a_{k+1} \ge a_{m+k+1} \kern-0.8pt=\kern-0.8pt \ell_{k+1}$,
$u_{k+1} \kern-0.8pt=\kern-0.8pt a_{k+1} \ge a_k \kern-0.8pt=\kern-0.8pt \ell_k$, we obtain
$\max_{i=1}^m l_i \le \min_{i=1}^m u_i$.
Figure~\ref{fig:bitonic_merge}d~shows that the two output sequences are bitonic and that all elements of the upper output sequence are greater than or equal to all elements of the lower output sequence.

\section{Implementation Details}

\subsection{MNIST}

For the MNIST based task, we use the same convolutional neural network architecture as in previous works \cite{Grover2019-NeuralSort, Cuturi2019-SortingOT}.
That is, two convolutional layers with a kernel size of $5\times5$, $32$ and $64$ channels respectively, each followed by a ReLU and MaxPool layer; 
after flattening, this is followed by a fully connected layer with a size of $64$, a ReLU layer, and a fully connected output layer mapping to a scalar.

\begin{figure*}
\vspace*{-2.5em}
\begin{multicols}{2}
\subsection{SVHN}
For the SVHN task, we use a network with four convolutional layers with a kernel size of $5\times5$ and ($32, 64, 128, 256$) filters, each followed by a ReLU and a max-pooling layer with stride $2\times2$; followed by a fully connected layer with size $64$, a ReLU, and a layer with output size $1$.

\subsection{Fast Sort \& Rank}
To evaluate the fast sorting and ranking method by \citet{Blondel2020-FastSorting}, we used the mean-squared-error loss between predicted and ground truth ranks as this method does not produce differentiable permutation matrices.

\subsection{Top-$k$ Supervision}
\label{sm:topk}
For top-$k$ supervision, we use ResNet18 as well as a Vanilla CNN with 4 convolutional and 2 fully connected layers.
The vanilla CNN is has the following architecture: \texttt{C16-BN-R-C32-BN-R-Max2-C64-BN-R-C128-BN} \texttt{-R-Max2-F256-Fc} where \texttt{Ck} denotes a convolutional layer with \texttt{k} output channels, a $3\times3$ kernel, and padding of $1$, \texttt{BN} denotes BatchNorm \cite{ioffe2015batch}, \texttt{R} denotes ReLU, \texttt{Max2} denotes MaxPool with a $2\times2$ kernel, and \texttt{Fk} denotes a fully connected layer with \texttt{k} outputs.
This vanilla CNN is inspired from \citet{Blondel2020-FastSorting}.
We train each model using Adam \cite{Kingma2014AdamOpt} for $500$ epochs at a learning rate of $10^{-3}$.

\vspace{1em}
\section{Standard Deviations of the Results}
\label{sm:std}
Tables~\ref{table:results-mnist-std}, \ref{table:results-svhn-std}, \ref{table:results-bitonic-large-std}, \ref{tab:ablation-study-std}, and~\ref{tab:top-k-std} display the standard deviations for the results in this work.

\end{multicols}
\end{figure*}

\begin{table*}[h]
    \caption{
    Same as Table~\ref{table:results-mnist} but with additional standard deviations.
    }
    \label{table:results-mnist-std}
    \centering
    \resizebox{1.\textwidth}{!}{
    \begin{tabular}{lccccc}
        \toprule
        \textbf{MNIST} & $n=3$ & $n=5$ & $n=7$ & $n=8$ & $n=15$  \\
        \midrule
        Fast Sort \& Rank  &  $90.6\ |\ 93.5\ |\ 73.5$  &  $71.5\ |\ 87.2\ |\ 71.5$  &  $49.7\ |\ 81.3\ |\ 70.5$  &  $29.0\ |\ 75.2\ |\ 69.2$  &  $2.8\ |\ 60.9\ |\ 67.4$ \\
          &  $\pm0.4\ |\ \pm0.3\ |\ \pm0.8$  &  $\pm0.9\ |\ \pm0.4\ |\ \pm0.9$  &  $\pm0.6\ |\ \pm0.3\ |\ \pm0.4$  &  $\pm1.1\ |\ \pm0.6\ |\ \pm0.7$  &  $\pm0.2\ |\ \pm0.4\ |\ \pm0.6$ \\
        \midrule            
        Odd-Even& $95.2\ |\ 96.7\ |\ 86.1$ & $86.3\ |\ 93.8\ |\ 86.3$ & $75.4\ |\ 91.2\ |\ 86.4$ & $64.3\ |\ 89.0\ |\ 86.7$ & $35.4\ |\ 83.7\ |\ 87.6$  \\
        & $\pm0.3\ |\ \pm0.2\ |\ \pm0.6$ & $\pm0.9\ |\ \pm0.4\ |\ \pm0.9$ & $\pm1.8\ |\ \pm0.6\ |\ \pm0.9$ & $\pm1.8\ |\ \pm0.6\ |\ \pm1.1$ & $\pm1.8\ |\ \pm0.5\ |\ \pm0.5$  \\
        \bottomrule \toprule
        \textbf{MNIST} & $n=2$ & $n=4$ & $n=8$ & $n=16$ & $n=32$  \\
        \midrule            
        Odd-Even& $98.1\ |\ 98.1\ |\ 84.3$ & $90.5\ |\ 94.9\ |\ 85.5$ & $63.6\ |\ 87.9\ |\ 83.6$ & $31.7\ |\ 82.8\ |\ 87.3$ & $1.7\ |\ 69.1\ |\ 86.7$  \\
        & $\pm0.3\ |\ \pm0.3\ |\ \pm0.9$ & $\pm1.2\ |\ \pm0.6\ |\ \pm1.5$ & $\pm11.6\ |\ \pm4.2\ |\ \pm6.1$ & $\pm1.5\ |\ \pm0.5\ |\ \pm0.5$ & $\pm0.5\ |\ \pm1.5\ |\ \pm1.0$  \\
        \midrule
        Bitonic      & $98.1\ |\ 98.1\ |\ {84.0}$ & $91.4\ |\ 95.3\ |\ 86.7$ & $70.6\ |\ 90.3\ |\ {86.9}$ & $30.5\ |\ 81.7\ |\ 86.6$ & $2.7\ |\ 67.3\ |\ 85.4$  \\
        & $\pm0.2\ |\ \pm0.2\ |\ \pm1.2$ & $\pm0.6\ |\ \pm0.3\ |\ \pm0.4$ & $\pm4.4\ |\ \pm1.3\ |\ \pm1.8$ & $\pm1.8\ |\ \pm1.2\ |\ \pm0.9$ & $\pm1.3\ |\ \pm2.7\ |\ \pm1.7$  \\
        \bottomrule
    \end{tabular}%
    }
\end{table*}

\begin{table*}[h]
    \caption{
    Same as Table~\ref{tab:svhn} but with additional standard deviations.
    }
    \label{table:results-svhn-std}
    \centering
    \resizebox{1.\textwidth}{!}{
    \begin{tabular}{lccccc}
        \toprule      
        \textbf{SVHN} & $n=2$ & $n=4$ & $n=8$ & $n=16$ & $n=32$  \\
        \midrule
        Det.~NeuralSort  &  $90.1\ |\ 90.1\ |\ 39.9$  &  $61.4\ |\ 78.1\ |\ 45.4$  &  $15.7\ |\ 62.3\ |\ 48.5$  &  $0.1\ |\ 45.7\ |\ 51.0$  &  $0.0\ |\ 29.9\ |\ 52.7$ \\
          &  $\pm0.7\ |\ \pm0.7\ |\ \pm1.7$  &  $\pm0.8\ |\ \pm0.3\ |\ \pm1.2$  &  $\pm1.6\ |\ \pm1.2\ |\ \pm1.6$  &  $\pm0.1\ |\ \pm0.6\ |\ \pm1.2$  &  $\pm0.0\ |\ \pm1.4\ |\ \pm1.5$ \\
        \midrule
        Optimal Transport  &  $85.5\ |\ 85.5\ |\ 25.9$  &  $57.6\ |\ 75.6\ |\ 41.6$  &  $19.9\ |\ 64.5\ |\ 51.7$  &  $0.3\ |\ 47.7\ |\ 53.8$  &  $0.0\ |\ 29.4\ |\ 53.3$ \\
          &  $\pm0.0\ |\ \pm0.0\ |\ \pm0.0$  &  $\pm1.1\ |\ \pm0.8\ |\ \pm1.8$  &  $\pm1.9\ |\ \pm1.1\ |\ \pm1.2$  &  $\pm0.2\ |\ \pm1.7\ |\ \pm1.4$  &  $\pm0.0\ |\ \pm1.0\ |\ \pm1.9$ \\
        \midrule            
        Fast Sort \& Rank   &  $93.4\ |\ 93.4\ |\ 57.6$  &  $58.0\ |\ 75.8\ |\ 41.5$  &  $8.6\ |\ 52.7\ |\ 34.4$  &   $0.3\ |\ 36.5\ |\ 41.6$  &  $0.0\ |\ 14.0\ |\ 27.5$ \\
         &  $\pm0.7\ |\ \pm0.7\ |\ \pm3.7$  &  $\pm1.1\ |\ \pm0.7\ |\ \pm1.0$  &  $\pm1.0\ |\ \pm0.6\ |\ \pm0.3$  & $\pm0.2\ |\ \pm1.4\ |\ \pm1.8$  &  $\pm0.0\ |\ \pm3.1\ |\ \pm9.1$ \\
        \midrule
        Odd-Even & $93.4\ |\ 93.4\ |\ 58.0$ & $74.8\ |\ 85.5\ |\ 62.6$ & $35.2\ |\ 73.5\ |\ 63.9$ & $1.8\ |\ 54.4\ |\ 62.3$ & $0.0\ |\ 36.6\ |\ 62.6$  \\
        & $\pm0.4\ |\ \pm0.4\ |\ \pm2.0$ & $\pm1.2\ |\ \pm0.7\ |\ \pm1.1$ & $\pm1.2\ |\ \pm0.5\ |\ \pm1.1$ & $\pm0.8\ |\ \pm1.6\ |\ \pm1.6$ & $\pm0.0\ |\ \pm1.5\ |\ \pm0.8$  \\
        \midrule
        Bitonic & $93.8\ |\ 93.8\ |\ 58.6$ & $74.4\ |\ 85.3\ |\ 62.1$ & $38.3\ |\ 75.1\ |\ 66.8$ & $3.9\ |\ 59.6\ |\ 66.8$ & $0.0\ |\ 42.4\ |\ 67.7$  \\
        & $\pm0.3\ |\ \pm0.3\ |\ \pm0.8$ & $\pm0.7\ |\ \pm0.3\ |\ \pm1.1$ & $\pm2.4\ |\ \pm1.1\ |\ \pm1.4$ & $\pm0.3\ |\ \pm0.8\ |\ \pm1.4$ & $\pm0.0\ |\ \pm3.5\ |\ \pm3.6$  \\
        \bottomrule
    \end{tabular}%
    }
\end{table*}

\begin{table*}[h]
    \caption{
    Same as Table~\ref{table:results-bitonic-large} but with additional standard deviations.
    }
    
    \label{table:results-bitonic-large-std}
    \centering
    \footnotesize
    \resizebox{1.\textwidth}{!}{
    \begin{tabular}{l|c|cccccc|cccccc}
        \toprule
        $\qquad\ARTstrength$      & $0.25$    & $0.4$     & $0.4$     & $0.4$     & $0.4$     & $0.4$     & $0.4$     & $0.4$     & $0.4$     & $0.4$     & $0.4$     & $0.4$     & $0.4$  \\
        \midrule
        $\qquad n$          & $32$      & $32$      & $64$      & $128$     & $256$     & $512$     & $1024$    & $32$      & $64$      & $128$     & $256$     & $512$     & $1024$  \\
        \midrule
        batch size          & $128$     & $128$     & $64$      & $32$      & $16$      & $8$       & $4$       & $4$       & $4$       & $4$       & $4$       & $4$       & $4$  \\
        \midrule
            $\steepness=30$ & $78.20$ & $79.89$ & $81.25$ & $82.50$ & $82.05$ & $82.50$ & $82.80$ & $71.08$ & $75.88$ & $79.43$ & $81.46$ & $82.98$ & $82.80$  \\
                            & $\pm2.35$ & $\pm1.97$ & $\pm1.93$ & $\pm1.09$ & $\pm2.62$ & $\pm1.75$ & $\pm2.27$ & $\pm1.67$ & $\pm2.30$ & $\pm2.35$ & $\pm1.47$ & $\pm2.02$ & $\pm2.27$  \\ \midrule
            $\steepness=32.5$ & $76.98$ & $79.62$ & $81.66$ & $80.15$ & $81.87$ & $82.64$ & $81.63$ & $72.31$ & $75.59$ & $79.71$ & $81.36$ & $82.99$ & $81.63$  \\
                            & $\pm0.86$ & $\pm3.62$ & $\pm2.42$ & $\pm3.84$ & $\pm2.19$ & $\pm1.60$ & $\pm6.22$ & $\pm2.04$ & $\pm2.05$ & $\pm1.57$ & $\pm1.98$ & $\pm1.67$ & $\pm6.22$  \\ \midrule
            $\steepness=35$ & $77.45$ & $80.93$ & $81.26$ & $80.72$ & $81.42$ & $81.51$ & $81.15$ & $71.15$ & $75.73$ & $78.81$ & $79.32$ & $82.30$ & $81.15$  \\
                            & $\pm1.64$ & $\pm2.75$ & $\pm2.41$ & $\pm3.89$ & $\pm2.09$ & $\pm2.12$ & $\pm3.12$ & $\pm1.69$ & $\pm2.46$ & $\pm1.36$ & $\pm4.85$ & $\pm1.22$ & $\pm3.12$  \\ \midrule
            $\steepness=37.5$ & $76.40$ & $80.02$ & $80.05$ & $81.50$ & $80.05$ & $82.67$ & $80.07$ & $70.69$ & $75.80$ & $79.11$ & $80.64$ & $82.70$ & $80.07$  \\
                            & $\pm3.90$ & $\pm1.74$ & $\pm1.93$ & $\pm2.03$ & $\pm3.94$ & $\pm2.21$ & $\pm3.67$ & $\pm2.26$ & $\pm1.22$ & $\pm1.88$ & $\pm2.18$ & $\pm1.66$ & $\pm3.67$  \\ \midrule
            $\steepness=40$ & $77.69$ & $80.97$ & $80.23$ & $81.55$ & $79.75$ & $81.89$ & $81.15$ & $70.20$ & $74.67$ & $78.14$ & $80.06$ & $81.39$ & $81.15$  \\
                            & $\pm1.54$ & $\pm2.03$ & $\pm3.51$ & $\pm1.97$ & $\pm5.41$ & $\pm2.51$ & $\pm3.31$ & $\pm2.06$ & $\pm2.45$ & $\pm2.49$ & $\pm1.93$ & $\pm1.67$ & $\pm3.31$  \\ \midrule
            mean            & $77.35$ & $80.29$ & $80.89$ & $81.28$ & $81.03$ & $82.24$ & $81.36$ & $71.09$ & $75.53$ & $79.04$ & $80.57$ & $82.47$ & $81.36$  \\
                            & $\pm2.06$ & $\pm2.48$ & $\pm2.48$ & $\pm2.80$ & $\pm3.48$ & $\pm2.03$ & $\pm3.97$ & $\pm2.00$ & $\pm2.10$ & $\pm1.97$ & $\pm2.77$ & $\pm1.71$ & $\pm3.97$  \\ \midrule
            best $s$        & $78.20$ & $80.97$ & $81.66$ & $82.50$ & $82.05$ & $82.67$ & $82.80$ & $72.31$ & $75.88$ & $79.71$ & $81.46$ & $82.99$ & $82.80$  \\
                            & $\pm3.90$ & $\pm3.62$ & $\pm3.51$ & $\pm3.89$ & $\pm5.41$ & $\pm2.51$ & $\pm6.22$ & $\pm2.26$ & $\pm2.46$ & $\pm2.49$ & $\pm4.85$ & $\pm2.02$ & $\pm6.22$  \\ \midrule
            worst $s$       & $76.40$ & $79.62$ & $80.05$ & $80.15$ & $79.75$ & $81.51$ & $80.07$ & $70.20$ & $74.67$ & $78.14$ & $79.32$ & $81.39$ & $80.07$  \\
                            & $\pm0.86$ & $\pm1.74$ & $\pm1.93$ & $\pm1.09$ & $\pm2.09$ & $\pm1.60$ & $\pm2.27$ & $\pm1.67$ & $\pm1.22$ & $\pm1.36$ & $\pm1.47$ & $\pm1.22$ & $\pm2.27$  \\ 
        \bottomrule
    \end{tabular}%
    }
\end{table*}

\begin{table*}[h]
    \caption{
    Same as Table~\ref{tab:ablation-study} but with additional standard deviations.
    }
    \label{tab:ablation-study-std}
    \centering
    \small
    \begin{tabular}{lcccc}
        \toprule
        & \multicolumn{2}{c}{$n = 4$} & \multicolumn{2}{c}{$n = 32$} \\
        \cmidrule(l){2-3}\cmidrule(l){4-5}
        Setting~~/~~$\ARTstrength$ & $0$ & $0.25$ & $0$ & $0.25$ \\
        \midrule
        Odd-Even (MNIST)    & $94.5 \pm0.3$ & $94.9 \pm0.6$ & $61.5 \pm1.9$ & $69.1 \pm1.5$  \\
        Bitonic (MNIST)     & $93.6 \pm1.4$ & $95.3 \pm0.3$ & $62.8 \pm15.5$ & $67.3 \pm2.7$  \\
        \midrule
        Odd-Even (SVHN)     & $77.3 \pm1.0$ & $85.5 \pm0.7$ & $28.5 \pm2.7$ & $36.6 \pm1.5$  \\
        Bitonic (SVHN)      & $78.1 \pm0.2$ & $85.3 \pm0.3$ & $35.0 \pm0.8$ & $42.4 \pm3.5$  \\
        \bottomrule
    \end{tabular}
\end{table*}

\begin{table*}
    \caption{
    Same as Table~\ref{tab:top-k} but with additional standard deviations.
    }
    \label{tab:top-k-std}
    \centering
    \small
    \begin{tabular}{lcccc}
        \toprule
        Setting                         & Softmax CE & Diff.~Top-$k$ \\
        \midrule
        CIFAR-10, Vanilla CNN           & $87.2\%\pm0.2\%$   & $88.0\%\pm0.4\%$               \\
        CIFAR-10, ResNet18              & $91.0\%\pm0.3\%$   & $90.9\%\pm0.2\%$                 \\
        CIFAR-100, Vanilla CNN          & $58.2\%\pm0.3\%$   & $56.3\%\pm0.5\%$                 \\
        CIFAR-100, ResNet18             & $61.9\%\pm0.4\%$   & $63.3\%\pm0.6\%$                   \\
        \bottomrule
    \end{tabular}
\end{table*}

\fi

\end{document}